\documentclass{article} 
\usepackage{iclr2023_conference,times}

\usepackage{amsmath,amsfonts,bm}

\def\eqref#1{equation~\ref{#1}}

\def\1{\bm{1}}

\DeclareMathAlphabet{\mathsfit}{\encodingdefault}{\sfdefault}{m}{sl}
\SetMathAlphabet{\mathsfit}{bold}{\encodingdefault}{\sfdefault}{bx}{n}

\usepackage{hyperref}
\usepackage{url}

\usepackage{algorithm} 
\usepackage{algpseudocode} 
\usepackage{amsmath}
\usepackage{graphicx}
\usepackage{xcolor}

\title{Agent Performing Autonomous Stock Trading under Good and Bad Situations}

\author{Yunfei Luo \\
Manning College of Information and Computer Science\\
University of Massachusetts Amherst\\
Amherst, MA 01003, USA \\
\texttt{yunfeiluo@umass.edu} \\
\And
Zhangqi Duan \\
Manning College of Information and Computer Science\\
University of Massachusetts Amherst \\
Amherst, MA 01003, USA \\
\texttt{zduan@umass.edu}
}

\iclrfinalcopy 
\begin{document}

\maketitle

\begin{abstract}
Stock trading is one of the popular ways for financial management. However, the market and the environment of economy is unstable and usually not predictable. Furthermore, engaging in stock trading requires time and effort to analyze, create strategies, and make decisions. It would be convenient and effective if an agent could assist or even do the task of analyzing and modeling the past data and then generate a strategy for autonomous trading. Recently, reinforcement learning has been shown to be robust in various tasks that involve achieving a goal with a decision making strategy based on time-series data. In this project, we have developed a pipeline that simulates the stock trading environment and have trained an agent to automate the stock trading process with deep reinforcement learning methods, including deep Q-learning, deep SARSA, and the policy gradient method. We evaluate our platform during relatively good (before 2021) and bad (2021 - 2022) situations. The stocks we've evaluated on including Google, Apple, Tesla, Meta, Microsoft, and IBM. These stocks are among the popular ones, and the changes in trends are representative in terms of having good and bad situations. 
We showed that before 2021, the three reinforcement methods we have tried always provide promising profit returns with total annual rates around $70\%$ to $90\%$, while maintain a positive profit return after 2021 with total annual rates around $2\%$ to $7\%$. 
\end{abstract}

\section{Introduction}
Stock trading is one of the well-known ways of financial management and investment \cite{finance-manage}. It will bring benefit and convenience if we are able to train an agent to learn how to extract patterns of past market then make decisions on the task of stock trading. Firstly, such autonomous system could save much time being put on analyzing past data. Secondly, the agent could act as an assistant to construct strategies and decide whether to take actions in an ongoing process. 

With the advantage mentioned, now comes the question of \textit{how to train the stock trading agent, how to guide the agent to make valid decisions, and how we know that the system is stable?} We realized that modeling the task of stock trading could be formalized as a Markov Decision Process (MDP) \cite{mdp}. We intend to train the agent using reinforcement learning methods so that it could learn from simulating the trading interactions with past data in order to come up with final strategies that could maximize the profit. Our expectation of the usage of this autonomous system is to assist ordinary people like us in the financial management and investment with a conservative approach that could bring some profit in a relatively small scale but not ruin the market with a larger scale. 

In this project, we have focused both on the MDP environment and the algorithms for training the agent. We constructed each part of the MDP by ourselves, and all three algorithms, including deep Q learning \cite{deep-q-learning-analysis}, deep SARSA \cite{deep-sarsa, deep-sarsa-auto-nav, deep-sarsa-offload}, and the policy gradient method \cite{policy-grad, policy-grad-robot}, were implemented from scratch. We also present the empirical results of the performance of agents trained with different algorithms. We observe that these agents always provide promising profit returns in good situations (before 2021) with annual total rates around $70\%$ to $90\%$. Because the background of economy in the real world was pretty pessimistic after 2021 probably due to the effect of the pandemic \cite{covid-impact-cause, covid-impact-china, covid-impact-global}, the performance of these agents get worse while still maintain a positive profit return after 2021 with total annual rates around $2\%$ to $7\%$. 
The details of our implementation of each component of the MDP, the algorithms, and the empirical results will be further elaborated in the following sections. 

\section{Related Work}
Financial management and investment is beneficial for both general public and the market \cite{finance-manage}. Stock trading is a popular way of financial management and investment, because the data are publicly available for analysis and inspection. There are many great works such as \citet{macd, rsi, cci, adx} that summarize past patterns and demonstrate the trend with a single number. These methods helps people inspect the market more conveniently, and are useful assistant for making the trading strategy. 

Reinforcement learning has been shown to be robust in many tasks \cite{rl-apply, rl-intro}. Stock trading can be formalized as a Markov Decision Process (MDP) \cite{mdp} that contains time series data from the environment with a decision-making policy to interact with the environment. Popular reinforcement learning methods including Q-Learning and SARSA \cite{q-learn, q-learn-sarsa-finance} are being applied to find the best policies in MDP. However, the task of stock trading is challenging for the vanilla version of these training methods. The main reason is that the data are in a continuous space, and discretizing the data could lead to a large number of parameters due to the large range of stock price. Another problem is that the discretizing approach could result in lack of generalization. Deep reinforcement learning methods \cite{deep-q-learning-analysis, deep-sarsa-offload, deep-sarsa, deep-sarsa-auto-nav, policy-grad, policy-grad-robot} have been shown that they are able to train the agent to perform the task under more complex environments \cite{deep-rl-apply, deep-rl-apply-power-system} by introducing deep neural network as part of the MDP then use gradient based method to optimize the parameters. \citet{deep_rl_for_st, deel_rl_for_st1, deel_rl_for_st2} have applied such a deep reinforcement learning method on the task of stock trading specifically, and they provide great guidelines for building the simulation pipeline and the environment for training the agent. Based on their clear presentations of their works, we are able to reproduce the entire environment and training pipeline efficiently for conducting further analysis and generalizing the technique. 

Recently, due to the pandemic, the stock market have been influenced greatly and become less stable \cite{covid-impact-cause, covid-impact-china, covid-impact-global}. For ordinary people like us would need to put more effort and time in analyzing the market and have to be more careful when making decisions. We consider the stock market during the pandemic to be a bad situation and the market during a relatively stable period to be a good situation. In this project, we train the agent using deep reinforcement learning algorithms and evaluate the performance in both good and bad situations. More details and insights will be presented in the following sections. 

\section{Methods}
\subsection{Environment}
We follow \cite{deep_rl_for_st} to configure the MDP environment for this stock trading task. The programming language that we used is Python, and each element of this MDP is presented in detail in successive subsections. 

\subsubsection{Input Data}
In this application, the input is time-series data indicating the price of a single share of stock, with time interval of 1 day. More details of the training and testing dates intervals will be provided in the Result \ref{sec:results} section. 

\subsubsection{Action}
The action space of the trading agent is whether to buy, hold, or sell the stock in each time step. We set a constraint here where the agent can trade at most $k$ stocks each time, either buy or sell. With $k > 0 \in \mathbb{Z}$ , we define the actions $A$ quantitatively as 
$$A = \{-k, -k+1, ..., -1, 0, 1, ..., k-1, k\}$$
where negative and positive values represent sales and purchases, respectively, and $0$ means holding. 

\subsubsection{State}
In this application, we essentially model the task of trading stocks as a Markov Decision Process (MDP). We initially define the state $s$ at time $t$ as 
$$s_t = [p_t, b_t, h_t]$$
where $p_t$ is the current price of a single share of stock, $b_t$ is the current balance available in the portfolio, and $h_t$ is the current number of shares hold in the portfolio. In addition, we introduce 4 market index as features in state as proposed in \cite{deep_rl_for_st}: Moving Average Convergence Divergence ($MACD$) \cite{macd}, Relative Strength Index ($RSI$) \cite{rsi},  Commodity Channel Index ($CCI$) \cite{cci}, Average Directional Index ($ADX$) \cite{adx}. These indices are hand made features where each of them summarize the trend in the past 2 or 3 weeks that are informative for decision making. Thus, our final version of the state is represented as
$$s_t = [p_t, b_t, h_t, MACD, RSI, CCI, ADX]$$
The transition from state $t$ to $t+1$ is deterministic. The price at $s_{t+1}$ is fixed in the training set, balances and number of shares are calculated based on the action $a_t$, and those market indices are recalculated based on the past data till $t+1$. 

\subsubsection{Reward}
The reward we set is the changing in the total asset value of the portfolio. More specifically, the reward $R$ given the states at time $t$ and $t+1$ with action $a_t$ is defined as
$$R(s_t, a_t, s_{t+1}) = (b_{t+1} + p_{t+1} \cdot h_{t+1}) - (b_t + p_t \cdot h_t) - c_t$$
where $c_t$ is the small fees applied on each trading event. This reward can be seen as the intermediate gain or loss of a portfolio after a trading event. 

\subsection{Assumption and Constraints}
We follow \cite{deep_rl_for_st} to construct the environment with the following assumptions:

\begin{itemize}
    \item Orders can be executed quickly at the close price.
    \item We assume that the stock market will not be affected by our reinforcement trading agent.
    \item Because the trading fees vary across platforms and services, for simplicity, we assume that the fee for each trade is $c_t = 0.1\% \cdot p_t \cdot (|h_t - h_{t-1}|)$, where $|h_t - h_{t-1}|$ represents the number of shares that are bought or sold. 
\end{itemize}

There are two constraints in the environment:
\begin{itemize}
    \item When conduct buying action, the total price cannot make the balance below the certain tolerance value (e.g. $\$0$, $\$-100$, etc). 
    \item When conducting a sale action, the number of shares being sold cannot exceed the number of shares that remain in the portfolio. 
\end{itemize}

\subsection{Deep Q-Learning}
The first algorithm we applied to train the agent is Q-Learning. Because we have the state declared as $s_t \in \mathbb{R}^7$ where $|S| = \infty$ and we have action with $|A| = 2k+1$, we cannot construct a table of $q(s, a)$ that contain all the combinations. By decretizing the state, we can form the table, but the space of $\mathbb{R}^7$ would make the table too large, hence still not feasible for applying the original Q-Learning algorithm. To address this situation, we train a neural network to approximate the $q$ function. The procedure of training the agent with deep Q-Learning is shown in algorithm \ref{algo:deep-q}, with $\alpha$ as learning rate, $\gamma$ as discount factor, $\pi$ as policy, and $q$ as the neural network that take state $s$ as input and output the $q$ values for each action (p.s. $q(s, a)$ means feed $s$ to $q$, and extract the value at corresponds to the index of $a$ in the output). 

\begin{algorithm}
	\caption{Deep Q-Learning, Input: $\alpha$, $\gamma$, $\pi$, $q$}
	\begin{algorithmic}[1]
            \State Initialize $\pi$ as uniformly distributed over actions for all states
		\For {$episode=1,2,\ldots$}
                \State Generate a trajectory from current policy $\pi$
                \State $\hat{q}(s, a) = COPY(q(s, a))$
                \For {$epoch=1, 2, \ldots$}
    			\For {Each $s_t, a_t, R_t$}
                        \State $q_{new}(s_t, a_t) = q(s_t, a_t) + \alpha(R_t + \gamma \max_{a'}\hat{q}(s_{t+1}, a') - q(s_t, a_t))$
                        \State Loss = HuberLoss$(q(s_t, a_t), q_{new}(s_t, a_t))$
                        \State Conduct Backpropagation from Loss
                        \State Update weights of $q$ with Adam Optimizer
    			\EndFor
                \EndFor
                \State Update $\pi$ with $\epsilon$-greedy strategy
		\EndFor
	\end{algorithmic} 
        \label{algo:deep-q}
\end{algorithm}

\begin{figure}[ht]
\begin{center}
\includegraphics[width=1.0\linewidth]{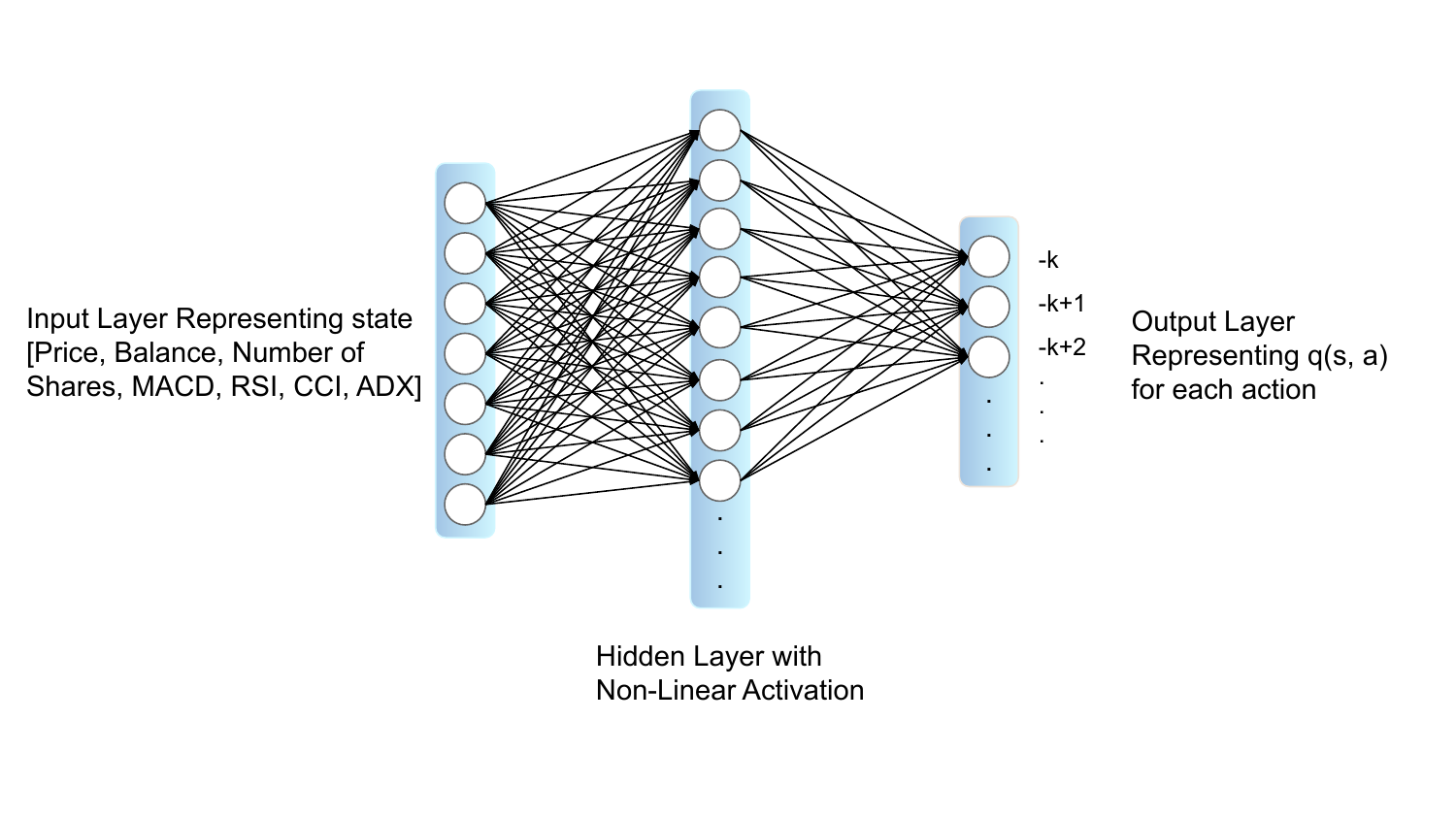}
\end{center}
\caption{Deep Q Network}
\label{fig:deep-q-network}
\end{figure}

More details of the algorithms are described below. First, the policy for generating the trajectory in each episode follows the $\epsilon$-greedy policies. The estimates optimal action is $a^* = argmax_{a' \in A}q(s, a')$. Then the actions in each step are sampled according to the following probabilities:
\[ \pi(s, a) = \begin{cases} 
      1 - \epsilon + \frac{\epsilon}{|A|} & \text{if } a = a^*\\
      \frac{\epsilon}{|A|} & \text{otherwise}
   \end{cases}
\]

Secondly, the loss function we choose as the objective function for $q$ neural networks is Huber Loss \cite{huber_loss, huber_loss_1} that have been shown to be robust for modeling regression tasks, because it kept the strength of both Mean Absolute Error (MAE) and Mean Squared Error (MSE). And the optimizer we use to update the weights of the neural network is Adam \cite{adam}. 

Finally, the neural network we construct for estimating the $q$ function is a three layers multi-layer perceptrons, with ReLU as activation function to create non-linearity. The workflow of this deep neural network is shown in Figure \ref{fig:deep-q-network}. 

\subsection{Deep SARSA}
Then, we implemented the Deep SARSA algorithm, which works in a similar way as Deep Q-Learning. We first tried the same neural network structure with Deep Q-Learning and then we have explored some other structures to find the model with best performance. It turns out that the same structure could generate the best performance. The difference between Deep Q-Learning and Deep SARSA is that instead of selecting update $q_{new}(s_t, a_t) = q(s_t, a_t) + \alpha(R_t + \gamma \max_{a'}\hat{q}(s_{t+1}, a') - q(s_t, a_t))$ and then compute the Huber loss and backpropagate to update the gradient, we found the next state and corresponding action from the simulation of the episode and then the upodate rule is: $q_{new}(s_t, a_t) = q(s_t, a_t) + \alpha(R_t + \gamma \hat{q}(s_{t+1}, a_{t+1}) - q(s_t, a_t))$. The Deep SARSA algorithm has the same hyperparameters as Deep Q-learning, but has different values after the hypertune process. In addition, the $\epsilon$-policies update rule has been kept in the Deep SARSA model. 

\subsection{Policy Gradient Method}
Furthermore, the third algorithm we implemented is the Policy Gradient Method. Instead of learning the parameters of the q function, the policy gradient method trained the parameters for the policy itself. To train the policy, we have updated the neural network model so that the last output will be a probability distribution over all action through the softmax function. The procedure of training the policy gradient model is shown in Algorithm 2. There are several parameters that are used as input in the algorithm, such as $\alpha$, which is the step size in updating the gradient, then $\gamma$ is used as the discount rate when calculating the return, and the policy $\pi$. 

\begin{algorithm}
	\caption{Policy Gradient, Input: $\alpha$, $\gamma$, $\pi$}
	\begin{algorithmic}[2]
            \State Initialize $\pi$ as uniformly distributed over actions for all states
		\For {$episode=1,2,\ldots$}
                \State Generate a trajectory from current policy $\pi$
                \For {$epoch=1, 2, \ldots$}
    			\For {Each step(t) in the episode}
                        \State $G_t = \sum_{K=t}^{T} \gamma ^{k-t}R_k$
                    \EndFor
                        \State Loss = $-\frac{1}{t} \sum_{t}^{T}ln(G_t \pi(a_t|s_t, \theta)$
                        \State Conduct Backpropagation from Loss
                        \State Update weights of $\pi$ with SGD Optimizer 
    			\EndFor
                \EndFor
	\end{algorithmic} 
        \label{algo:policy gradient}
\end{algorithm}

It is worth noticing that we actually take the negative sign when we calculated the loss during the policy gradient algorithm due to that this is an optimizing process instead of minimizing step. Then, by multiplying with -1, we could update the gradient with the right direction that we want to train the model parameters. Moreover, to optimize the hyper parameters, we have tested several sets of hyperparameters including the learning rate, gamma, epsilon, episodes, and epochs to obtain the model with best performance. It turns out that the hyperparameter setting actually keeps the same for three models and the difference is the choice of optimizer between the deep learning algorithms and policy gradient method.

\section{Results} \label{sec:results}
\subsection{Evaluation Schema}
We evaluate the reinforcement learning methods in two scenarios, where one is conducted before the beginning of 2021, another is conducted after 2021. Because, starting in 2021, many of the stocks undergo relatively very different trends than they had before 2021. The training and testing range is shown in Table \ref{tab:train-test} where the dates are presented with order: month, day, year. 

\begin{table}[ht]
    \centering
    \begin{tabular}{|c|c|c|}
        \hline
        Scenarios & Train & Test \\
        \hline\hline
        Before 2021 & 01-08-2013 to 01-02-2019 & 01-03-2019 to 12-30-2020\\
        After 2021 & 01-08-2013 to 12-08-2020 & 12-09-2020 to 12-01-2022\\
        \hline
    \end{tabular}
    \caption{Overview of data sets.}
    \label{tab:train-test}
\end{table}

\begin{figure}[ht]
\begin{center}
\includegraphics[width=1.2\linewidth]{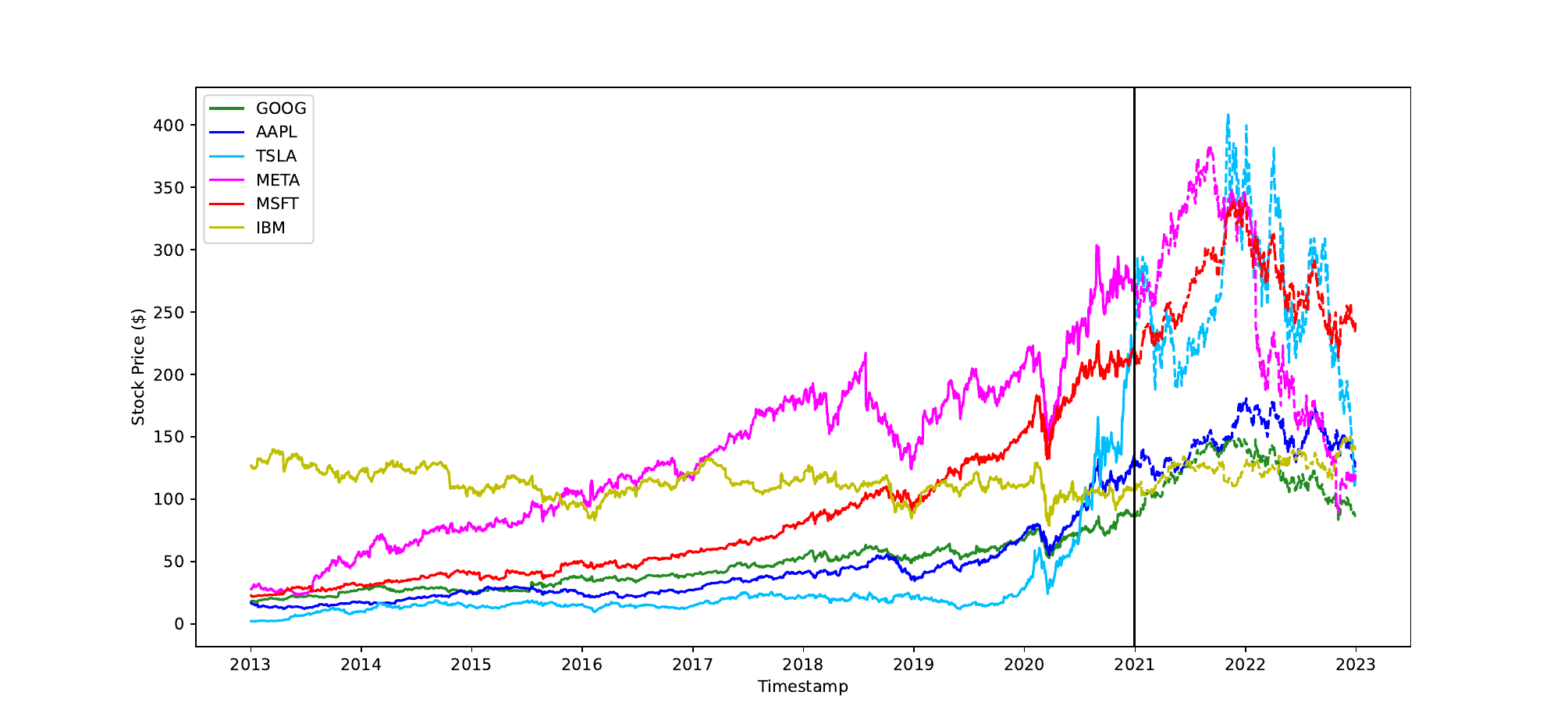}
\end{center}
\caption{Prices of the selected stocks. The split line separate the good and bad situations. }
\label{fig:data-trends}
\end{figure}

We choose to conduct training and testing on the stocks of six flag companies in the field of technology: Google, Apple, Tesla, Meta, Microsoft, and IBM. These stocks are among the popular ones, and the changes in trends are representative in terms of having good and bad situations. The good situation occurs when the market is relatively stable and the prices are increasing overall. The bad situation occurs when the prices oscillate with patterns very differently from the patterns of past data. The data we used are shown in Figure \ref{fig:data-trends}. 

\begin{figure}[ht]
\begin{center}
\includegraphics[width=0.8\linewidth]{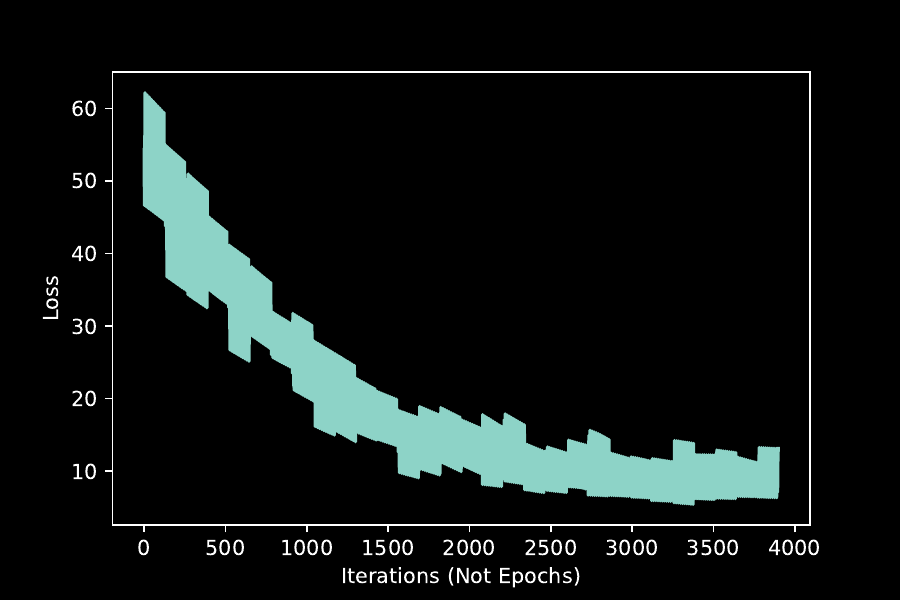}
\end{center}
\caption{Training loss of Q function}
\label{fig:loss-example}
\end{figure}

The initial balance for each stock is $\$1000$ (i.e. $\$6000$ in total for the investment), and each trading event (buy or sell) contains at most $5$ numbers of shares. There is a tolerance that the balance can go to negative to $\$-100$ for each stock. 
For hyperparameters, we tuned the setting until we observed exponentially decreasing trends in loss of estimation by neural networks, such as the one shown in Figure \ref{fig:loss-example}. The figure shows that the loss decreases during the training process and finally converges. The final setting is shown in Table \ref{tab:hyper-param} and applies to the three models we have implemented. 

\begin{table}[ht]
    \centering
    \begin{tabular}{|c|c|}
        \hline
        Hyperparameter & Value \\
        \hline\hline
        Episode & 30\\
        Discount Factor $\gamma$ & 0.6\\
        Learning Rate $\alpha$ & 0.7 \\
        Initial $\epsilon$ policy & 0.8 \\
        Minimum $\epsilon$ policy & 0.2 \\ 
        $\epsilon$ Decay & linearly Decay with factor of 0.9\\
        Epochs for Updating NN & 10 \\
        Learning Rate for NN & 1e-5 \\
        \hline
    \end{tabular}
    \caption{Setting of hyperparameters. The notations stay consistent as described in previous sections. NN stands for Neural Network}
    \label{tab:hyper-param}
\end{table}

As we have shown in the previous algorithm, the optimizer used for deep Q-Learning and deep SARSA is Adam, and the optimizer we selected eventually for policy gradient is SGD. 

Furthermore, we run the training and testing for a repetition of 20 times and report the average performance in the following subsections. The percentiles to show the variations are presented in the Appendix. 

\subsection{Before 2021 - Good Situation}
Table \ref{tab:before-2021} presents the profit along with the annual return rate on the test set before the beginning of the year 2021. 

\begin{table}[ht]
    \centering
    \begin{tabular}{|c|c|c|c|c|c|c|c|}
        \hline
        Method & GOOG & AAPL & TSLA & META & MSFT & IBM & Total\\
        \hline\hline
        Deep Q-Learning | Profit & \$366.86 & \bf \$1618.07 & \$7164.62 & \bf \$285.62 & \$458.37 & \$115.72 & \$10009.26\\
        Annual Rate & 18.343\% & \bf 80.9035\% & 358.231\% & \bf 14.281\% & 22.9185\% & 5.786\% & 83.4105\% \\\hline
        Deep SARSA | Profit &\$293.38  &\$1041.29 & \$6602.23 & \$211.81& \$218.98 & \bf \$125.98 & \$8493.67\\
        Annual Rate & 14.669\%  & 52.06\%&330.112\% & 10.59\%& 10.949\% & \bf 6.299\%& 70.7805\%\\\hline
        Policy Gradient | Profit & \bf \$472.72 &\$1268.24 & \bf \$8254.33 &\$228.21 & \bf \$633.64 &\$111.83 & \bf \$10969\\
        Annual Rate & \bf 23.63\% &63.41\% & \bf 412.717\% & 11.41\% & \bf 31.68\% &5.59\% & \bf 91.408\%\\\hline
    \end{tabular}
    \caption{Returned Profits on the test set before 2021. }
    \label{tab:before-2021}
\end{table}

In the first part which we used the data before 2021, all three models are trained to make decisions that could make a profit in each stock we chosen. In general, the three methods that we have tested could achieve competitive performance.
It seems that the models have learned useful information and converge based on our observation \ref{app-learning-curve}. In addition, we have also includes the percentiles of the performance over 20 repetition trails that demonstrate the variation of the experiments \ref{app-percentile}. 

We observe that the agents are able to maintain a positive return on average. These sets of experiments mainly evaluate the autonomous system we have developed in good situations, where the main trend of the stock market is increasing and relatively stable when we look at the past data. 

\subsection{Starting 2021 to 2022 - Bad Situation}

Table \ref{tab:after-2021} present the profit along with annual return rate on the test set after the beginning of the year 2021.


\begin{table}[ht]
    \centering
    \begin{tabular}{|c|c|c|c|c|c|c|c|}
        \hline
        Method & GOOG & AAPL & TSLA & META & MSFT & IBM & Total\\
        \hline\hline
        Deep Q-Learning | Profit & \$118.69 & \$115.78 & \bf \$3.73 & \$-271.93 & \$114.43 & \bf \$253.38 & \$334.08\\
        Annual Rate & 5.9345\% & 5.789\% & \bf 0.1865\% & -13.5965\% & 5.7215\% & \bf 12.669\% & 2.784\% \\\hline
        Deep SARSA | Profit & \bf \$143.08& \bf \$216.16 &\$-41.06 & \bf \$130.12 & \bf \$118.73&\$126.51 & \bf \$893.54 \\
        Annual Rate & \bf 7.154\% & \bf 10.8\% & -2.053\%& \bf 6.51\% & \bf 5.93\% & 6.3255\% & \bf 7.4461\%\\\hline
        Policy Gradient | Profit &\$67.75 & \$176.09 &\$-16.48 & \$-297.69&\$114.73 & \$220.81 &\$265.21 \\
        Annual Rate &3.39\% &8.804\% &-0.824\% &-14.88\% & 5.74\%& 11.04\%& 2.21\%\\\hline
    \end{tabular}
    \caption{Returned Profits on the test set after 2021. }
    \label{tab:after-2021}
\end{table}

Then, when we used the data after 2021 to train the model, the performance of all models in each stock decreased. There are cases that the models are actually losing money at the end of the training. The performance of the neural network is affected not only by the structure and hyperparameters, but also by the input data. Because the data distribution of test set from 2021 to 2022 is very different from the data distribution before 2021, the neural networks we've embedded in the agents is lack of generalization on the trends that they never seen during the training. 

In this bad situation, although the stock trading agents could achieve a positive return in the end, there are more occurrence of cases with negative returns compared to the performance in the test set in a good situation. As we have discussed, the shift in the distribution of data is the main causation of the degrade in performance. Introducing data from the financial crisis into the training set could mitigate this degradation. We leave this further validation for potential improvements in returns during the bad situations in future work. 

\section{Conclusion}
In this project, we trained agents for autonomous stock trading with three representative deep reinforcement learning methods, including the deep Q-Learning, the deep SARSA, and the policy gradient method. We evaluate the methods under two scenarios, one is before the beginning of 2021 where the market is relative stable, another one is after 2021 where the market experience lots of variation with bad situations. We show that the agents we trained are able to obtain a considerable return on the test set before 2021, while still obtaining the main positive return on the test set after 2021. 
We intend to develop this autonomous system to assist ordinary people like us in engaging in financial management and investment with a conservative approach that could bring profit on a relatively small scale but not ruin the market on a larger scale. Future work would involve experimenting with a multitask or federated learning approach instead of training individual agents for all stocks. We would also like to introduce data like those in the financial crisis in 2008 to the training set to explore the potential improvements of the agent's performance under bad situations. 

\subsubsection*{Data Availability}
The data analyzed during the current study are available from: \href{https://pypi.org/project/yfinance/}{Yahoo Finance}.

\subsubsection*{Code Availability}
The custom codes used for conducting experiments and generating results during the current study are available in the repository on GitHub: \href{https://github.com/yunfeiluo/Autonomous-Stock-Trading}{https://github.com/yunfeiluo/Autonomous-Stock-Trading}. 

\subsubsection*{Acknowledgments}
We owe our deepest gratitude to Professor Bruno Castro da Silva for constructive feedback and patient guidance through the reinforcement learning course he had taught at the University of Massachusetts Amherst.

\bibliography{iclr2023_conference}

\begin{thebibliography}{27}
\providecommand{\natexlab}[1]{#1}
\providecommand{\url}[1]{\texttt{#1}}
\expandafter\ifx\csname urlstyle\endcsname\relax
  \providecommand{\doi}[1]{doi: #1}\else
  \providecommand{\doi}{doi: \begingroup \urlstyle{rm}\Url}\fi

\bibitem[Adrian(2011)]{rsi}
ran-Moroan Adrian.
\newblock The relative strength index revisited.
\newblock \emph{African Journal of Business Management}, 5\penalty0
  (14):\penalty0 5855--5862, 2011.

\bibitem[Aggarwal et~al.(2021)Aggarwal, Nawn, and Dugar]{covid-impact-cause}
Shobhit Aggarwal, Samarpan Nawn, and Amish Dugar.
\newblock What caused global stock market meltdown during the covid
  pandemic--lockdown stringency or investor panic?
\newblock \emph{Finance research letters}, 38:\penalty0 101827, 2021.

\bibitem[Aguirre et~al.(2020)Aguirre, Medina, and M{\'e}ndez]{macd}
Alberto Antonio~Agudelo Aguirre, Ricardo Alfredo~Rojas Medina, and N{\'e}stor
  Dar{\'\i}o~Duque M{\'e}ndez.
\newblock Machine learning applied in the stock market through the moving
  average convergence divergence (macd) indicator.
\newblock \emph{Investment Management \& Financial Innovations}, 17\penalty0
  (4):\penalty0 44, 2020.

\bibitem[Alfakih et~al.(2020)Alfakih, Hassan, Gumaei, Savaglio, and
  Fortino]{deep-sarsa-offload}
Taha Alfakih, Mohammad~Mehedi Hassan, Abdu Gumaei, Claudio Savaglio, and
  Giancarlo Fortino.
\newblock Task offloading and resource allocation for mobile edge computing by
  deep reinforcement learning based on sarsa.
\newblock \emph{IEEE Access}, 8:\penalty0 54074--54084, 2020.

\bibitem[Andrecut \& Ali(2001)Andrecut and Ali]{deep-sarsa-auto-nav}
M~Andrecut and MK~Ali.
\newblock Deep-sarsa: A reinforcement learning algorithm for autonomous
  navigation.
\newblock \emph{International Journal of Modern Physics C}, 12\penalty0
  (10):\penalty0 1513--1523, 2001.

\bibitem[Corazza \& Sangalli(2015)Corazza and Sangalli]{q-learn-sarsa-finance}
Marco Corazza and Andrea Sangalli.
\newblock Q-learning and sarsa: a comparison between two intelligent stochastic
  control approaches for financial trading.
\newblock \emph{University Ca'Foscari of Venice, Dept. of Economics Research
  Paper Series No}, 15, 2015.

\bibitem[Fabozzi \& Peterson(2003)Fabozzi and Peterson]{finance-manage}
Frank~J Fabozzi and Pamela~P Peterson.
\newblock \emph{Financial management and analysis}, volume 132.
\newblock John Wiley \& Sons, 2003.

\bibitem[Fan et~al.(2020)Fan, Wang, Xie, and Yang]{deep-q-learning-analysis}
Jianqing Fan, Zhaoran Wang, Yuchen Xie, and Zhuoran Yang.
\newblock A theoretical analysis of deep q-learning.
\newblock In \emph{Learning for Dynamics and Control}, pp.\  486--489. PMLR,
  2020.

\bibitem[Gurrib(2018)]{adx}
Ikhlaas Gurrib.
\newblock Performance of the average directional index as a market timing tool
  for the most actively traded usd based currency pairs.
\newblock \emph{Banks and Bank Systems}, 13\penalty0 (3):\penalty0 58--70,
  2018.

\bibitem[Hastie et~al.(2009)Hastie, Tibshirani, Friedman, and
  Friedman]{huber_loss_1}
Trevor Hastie, Robert Tibshirani, Jerome~H Friedman, and Jerome~H Friedman.
\newblock \emph{The elements of statistical learning: data mining, inference,
  and prediction}, volume~2.
\newblock Springer, 2009.

\bibitem[Huber(1964)]{huber_loss}
Peter~J. Huber.
\newblock {Robust Estimation of a Location Parameter}.
\newblock \emph{The Annals of Mathematical Statistics}, 35\penalty0
  (1):\penalty0 73 -- 101, 1964.
\newblock \doi{10.1214/aoms/1177703732}.
\newblock URL \url{https://doi.org/10.1214/aoms/1177703732}.

\bibitem[Kingma \& Ba(2014)Kingma and Ba]{adam}
Diederik~P Kingma and Jimmy Ba.
\newblock Adam: A method for stochastic optimization.
\newblock \emph{arXiv preprint arXiv:1412.6980}, 2014.

\bibitem[Lee et~al.(2021)Lee, Lee, and Wu]{covid-impact-china}
Chi-Chuan Lee, Chien-Chiang Lee, and Yizhong Wu.
\newblock The impact of covid-19 pandemic on hospitality stock returns in
  china.
\newblock \emph{International Journal of Finance \& Economics}, 2021.

\bibitem[Li(2017)]{deep-rl-apply}
Yuxi Li.
\newblock Deep reinforcement learning: An overview.
\newblock \emph{arXiv preprint arXiv:1701.07274}, 2017.

\bibitem[Liu et~al.(2022)Liu, Ventre, and Polukarov]{deel_rl_for_st2}
Chunli Liu, Carmine Ventre, and Maria Polukarov.
\newblock Synthetic data augmentation for deep reinforcement learning in
  financial trading.
\newblock In \emph{Proceedings of the Third ACM International Conference on AI
  in Finance}, pp.\  343--351, 2022.

\bibitem[Liu et~al.(2021)Liu, Yang, Gao, and Wang]{deel_rl_for_st1}
Xiao-Yang Liu, Hongyang Yang, Jiechao Gao, and Christina~Dan Wang.
\newblock Finrl: Deep reinforcement learning framework to automate trading in
  quantitative finance.
\newblock In \emph{Proceedings of the Second ACM International Conference on AI
  in Finance}, pp.\  1--9, 2021.

\bibitem[Maitah et~al.(2016)Maitah, Prochazka, Cermak, and
  {\v{S}}r{\'e}dl]{cci}
Mansoor Maitah, Petr Prochazka, Michal Cermak, and Karel {\v{S}}r{\'e}dl.
\newblock Commodity channel index: Evaluation of trading rule of agricultural
  commodities.
\newblock \emph{International Journal of Economics and Financial Issues},
  6\penalty0 (1):\penalty0 176--178, 2016.

\bibitem[Naeem et~al.(2020)Naeem, Rizvi, and Coronato]{rl-intro}
Muddasar Naeem, Syed Tahir~Hussain Rizvi, and Antonio Coronato.
\newblock A gentle introduction to reinforcement learning and its application
  in different fields.
\newblock \emph{IEEE access}, 8:\penalty0 209320--209344, 2020.

\bibitem[Ngwakwe et~al.(2020)]{covid-impact-global}
Collins~C Ngwakwe et~al.
\newblock Effect of covid-19 pandemic on global stock market values: a
  differential analysis.
\newblock \emph{Acta Universitatis Danubius. {\OE}conomica}, 16\penalty0
  (2):\penalty0 255--269, 2020.

\bibitem[Peters \& Schaal(2006)Peters and Schaal]{policy-grad-robot}
Jan Peters and Stefan Schaal.
\newblock Policy gradient methods for robotics.
\newblock In \emph{2006 IEEE/RSJ International Conference on Intelligent Robots
  and Systems}, pp.\  2219--2225. IEEE, 2006.

\bibitem[Puterman(1990)]{mdp}
Martin~L Puterman.
\newblock Markov decision processes.
\newblock \emph{Handbooks in operations research and management science},
  2:\penalty0 331--434, 1990.

\bibitem[Qiang \& Zhongli(2011)Qiang and Zhongli]{rl-apply}
Wang Qiang and Zhan Zhongli.
\newblock Reinforcement learning model, algorithms and its application.
\newblock In \emph{2011 International Conference on Mechatronic Science,
  Electric Engineering and Computer (MEC)}, pp.\  1143--1146. IEEE, 2011.

\bibitem[Sutton et~al.(1999)Sutton, McAllester, Singh, and
  Mansour]{policy-grad}
Richard~S Sutton, David McAllester, Satinder Singh, and Yishay Mansour.
\newblock Policy gradient methods for reinforcement learning with function
  approximation.
\newblock \emph{Advances in neural information processing systems}, 12, 1999.

\bibitem[Watkins \& Dayan(1992)Watkins and Dayan]{q-learn}
Christopher~JCH Watkins and Peter Dayan.
\newblock Q-learning.
\newblock \emph{Machine learning}, 8:\penalty0 279--292, 1992.

\bibitem[Yang et~al.(2020)Yang, Liu, Zhong, and Walid]{deep_rl_for_st}
Hongyang Yang, Xiao-Yang Liu, Shan Zhong, and Anwar Walid.
\newblock Deep reinforcement learning for automated stock trading: An ensemble
  strategy.
\newblock In \emph{Proceedings of the First ACM International Conference on AI
  in Finance}, pp.\  1--8, 2020.

\bibitem[Zhang et~al.(2019)Zhang, Zhang, and Qiu]{deep-rl-apply-power-system}
Zidong Zhang, Dongxia Zhang, and Robert~C Qiu.
\newblock Deep reinforcement learning for power system applications: An
  overview.
\newblock \emph{CSEE Journal of Power and Energy Systems}, 6\penalty0
  (1):\penalty0 213--225, 2019.

\bibitem[Zhao et~al.(2016)Zhao, Wang, Shao, and Zhu]{deep-sarsa}
Dongbin Zhao, Haitao Wang, Kun Shao, and Yuanheng Zhu.
\newblock Deep reinforcement learning with experience replay based on sarsa.
\newblock In \emph{2016 IEEE symposium series on computational intelligence
  (SSCI)}, pp.\  1--6. IEEE, 2016.

\end{thebibliography}
\bibliographystyle{iclr2023_conference}

\appendix
\section{Learning Curve} \label{app-learning-curve}
\begin{figure}[ht]
\begin{center}
\includegraphics[width=0.8\linewidth]{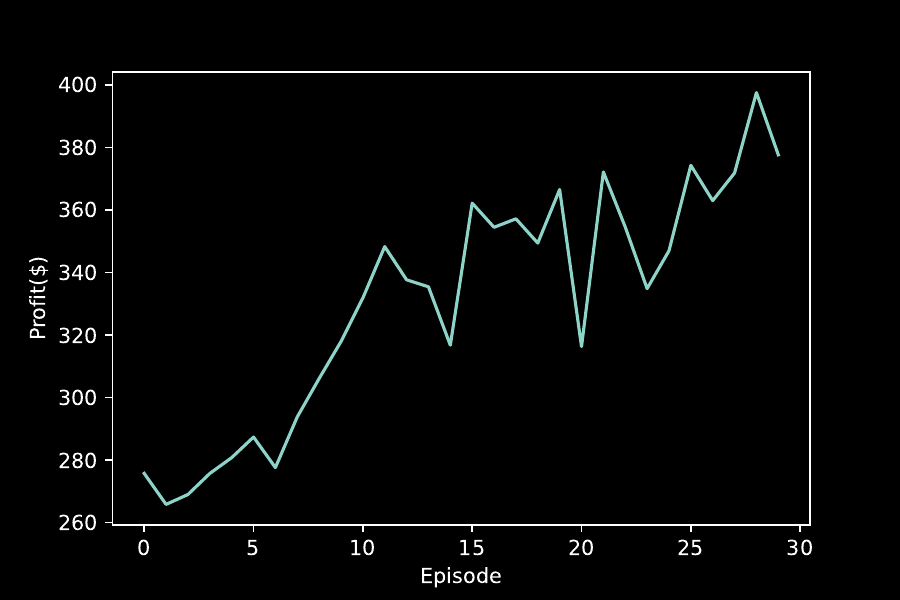}
\end{center}
\caption{Learning curve of performance in GOOG stock in the test set before 2021. }
\label{fig:lr-goog}
\end{figure}

\begin{figure}[ht]
\begin{center}
\includegraphics[width=0.8\linewidth]{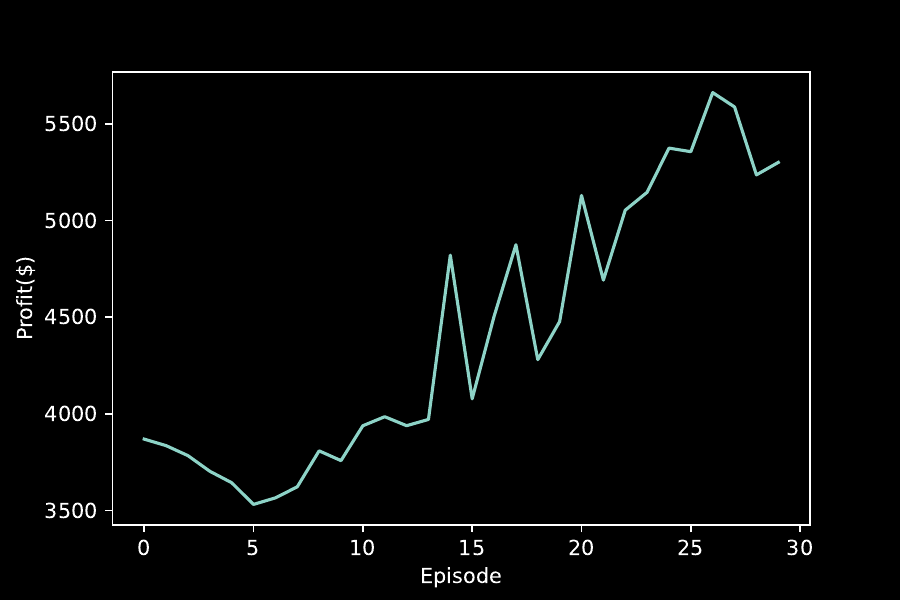}
\end{center}
\caption{Learning curve of performance in TSLA stock in the test set before 2021. }
\label{fig:lr-tsla}
\end{figure}

The following are several representative learning curve recorded during the training process using different models, where the agent progressively learn the trading strategy to maximize the profit. 
The performance (profit) vs. iteration figures are shown in \ref{fig:lr-goog} \ref{fig:lr-tsla}. We believe that the learning curve has shown that all models have converged during the process and have learned from the input data. 

\section{Percentiles} \label{app-percentile}
\begin{figure}[ht]
\begin{center}
\includegraphics[width=1.2\linewidth]{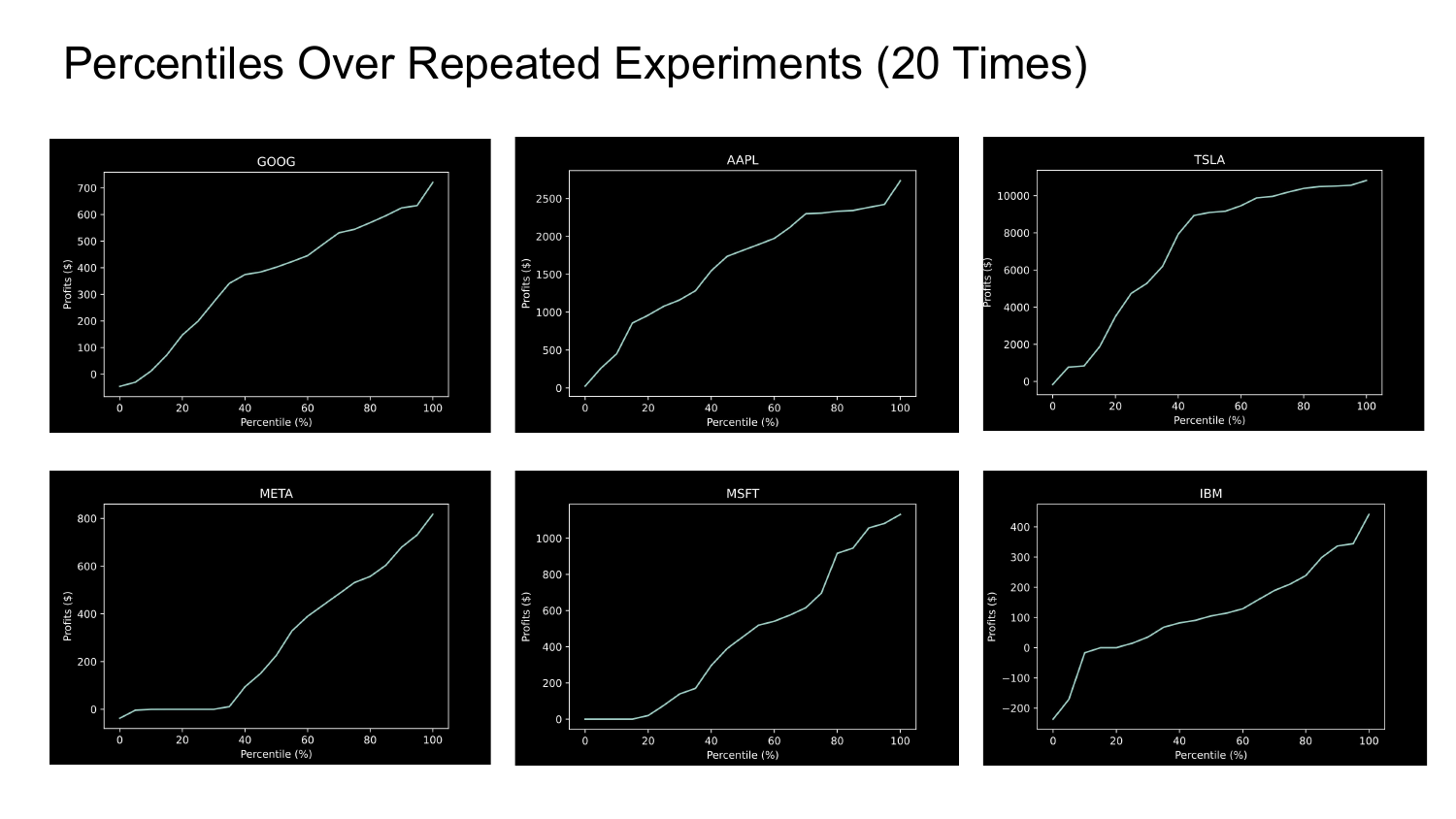}
\end{center}
\caption{}
\label{fig:percentile-before}
\end{figure}

\begin{figure}[ht]
\begin{center}
\includegraphics[width=1.2\linewidth]{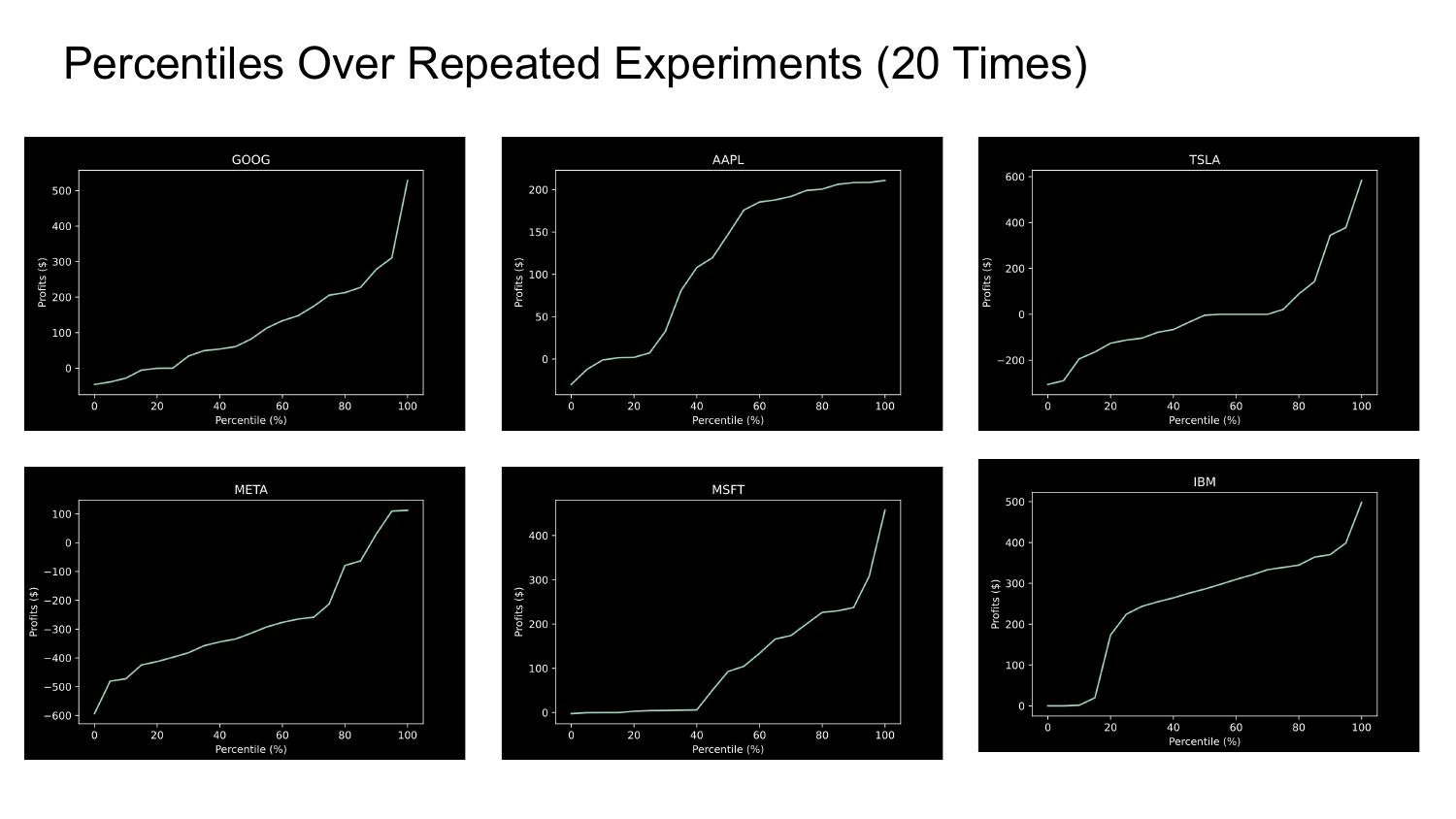}
\end{center}
\caption{}
\label{fig:percentile-after}
\end{figure}

We present the percentile of the performance over 20 time of repetition trails to show the variation of the experiments \ref{fig:percentile-before} \ref{fig:percentile-after}. 

\end{document}